\documentclass[11pt]{article}
\usepackage[margin=1in]{geometry}
\usepackage{graphicx}
\usepackage{booktabs}
\usepackage{siunitx}
\usepackage{amsmath}
\usepackage{amssymb}
\usepackage{bm}
\usepackage{float}
\usepackage{rotating}
\usepackage{algorithm}
\usepackage{algpseudocode}
\usepackage[numbers,sort&compress]{natbib}
\usepackage[hidelinks]{hyperref}
\usepackage[noblocks]{authblk}
\title{A Multi-Agent System for Autonomous, Fine-Tuning-Free Clinical Symptom Detection: Development and Validation Study}
\author[1]{Cameron Cagan}
\author[1]{Pedram Fard}
\author[1]{Jiazi Tian}
\author[1]{Jingya Cheng}
\author[2]{Shawn N. Murphy}
\author[1]{Hossein Estiri}

\affil[1]{Department of Medicine, Massachusetts General Hospital, Boston, MA, USA}
\affil[2]{Department of Biomedical Informatics and Medical Education, University of Washington, WA, USA}
\date{}
\begin{document}
\maketitle

\begin{abstract}
Clinical notes contain many of the signs and symptoms that bring patients to care, yet this information rarely reaches structured fields. Existing extraction approaches either rely on context-insensitive rules that generate false positives or on supervised models that require substantial fine-tuning. We present Pythia, a multi-agent system that autonomously writes and optimizes extraction prompts for clinical concepts without manual prompt engineering or fine-tuning. Running on a locally hosted open-weights model, Pythia keeps clinical notes on local infrastructure and selects prompts using development-set sensitivity and specificity. We compared Pythia with a curated lexicon across 72 signs and symptoms from 400 clinical notes representing 387 patients. Development (n=300) and validation (n=100) sets were partitioned independently for each concept. Pythia achieved mean sensitivity of 0.76 and specificity of 0.95, compared with 0.82 and 0.76 for the lexicon, and matched or exceeded the lexicon on both metrics for 20 of 62 directly comparable concepts. For 14 concepts where the lexicon labeled every note positive, Pythia recovered mean specificity of 0.97 by requiring a present-tense, patient-attributed finding rather than any textual mention of a term. Specificity transferred from development to validation with minimal degradation across prevalences, whereas sensitivity transfer weakened below 5\% prevalence, reaching a mean gap of 0.25 below 2\% prevalence. A BERT classifier fine-tuned per concept on the same development set achieved mean sensitivity of 0.23 and collapsed to zero sensitivity for concepts below roughly 5\% prevalence. These findings suggest that autonomous, fine-tuning-free prompt optimization can produce symptom extraction prompts that generalize effectively from development to validation while remaining deployable on local infrastructure.
\end{abstract}

\section{Introduction}

Clinical notes record the signs and symptoms that bring patients to care, and much of that information never reaches the structured fields of electronic health records. Diagnosis codes and laboratory values capture confirmed conditions, while the presenting complaints and the day-to-day course of illness live in free text. Studies that depend on symptoms, including computational phenotyping, surveillance, and the generation of real-world evidence, therefore require extraction of these findings from notes at scale \cite{hripcsak2013,banda2018,wang2018review}. The quality of that extraction sets a ceiling on every downstream use. Post-acute COVID-19 illustrates the problem, because its case definition rests on a broad and shifting set of signs and symptoms that are documented almost entirely in narrative text rather than in diagnosis codes.

Extracting a symptom reliably is more complex than just finding the symptom name in a piece of text. The same term often appears in negated statements, instructions to the patient, family history, and more. Separating a present finding from these other uses, through negation, attribution, and temporality, has been a central problem in clinical natural language processing for two decades \cite{chapman2001negex,wang2018review,sheikhalishahi2019}. Rule-based lexicons remain the default because they are transparent and require no labeled training data, yet they do not consider context, and their false-positive burden increases with the size of the negative class \cite{savova2010ctakes,soysal2018clamp}.

Learning-based methods provide an avenue for incorporating context. Transformer encoders trained on annotated notes reached strong extraction performance \cite{si2019contextual,alsentzer2019clinicalbert,gu2021pubmedbert}. Further, a recent multi-institution benchmark placed the current state of the art at fine-tuned large language models, which surpass encoders when labeled data is scarce \cite{hu2026ie,chen2025benchmark}. These gains carry a cost, as the benchmark models required thousands of annotated notes and weight fine-tuning, and they are far slower than the encoder baseline \cite{hu2026ie}. Prompting lowers the training cost but not the design cost, because few-shot and hand-engineered prompts still depend on a person to write and revise them \cite{agrawal2022fewshot,hu2024prompt}. Automatic prompt optimization searches instruction space against a metric without weight updates \cite{shin2020autoprompt,zhou2022ape,yang2023opro,khattab2023dspy}, and open-weights models now run inside the institution for privacy-sensitive clinical text \cite{wiest2025local,gptoss}. The binding constraint on building these systems is the labeled data and the manual prompt design they require rather than the model.

This shift has a name in agentic software engineering. Under loop engineering, the designer stops writing prompts for an agent and instead builds a loop that sets a goal, lets the agent iterate toward it, and uses a separate process to check whether the goal is met before stopping \cite{osmani2026loop}. The pattern rests on established machinery, because automatic prompt optimization already searches instruction space against a metric without weight updates \cite{shin2020autoprompt,zhou2022ape,yang2023opro,khattab2023dspy}, and multi-agent designs already separate the agent that proposes a solution from the one that verifies it. Clinical natural language processing has rarely evaluated such loops against the sensitivity and specificity requirements that clinical use imposes. Whether an autonomous prompt loop can meet those clinical requirements remains an open question.

We study Pythia, a system that instantiates loop engineering for clinical symptom detection and removes both the fine-tuning cost of encoder models and the per-concept design cost of manual prompting. Pythia writes and optimizes its own extraction prompt for each concept, with no manual prompt engineering and no weight fine-tuning, and it runs on an open-weights model so that notes stay on local infrastructure \cite{estiri_pythia, gptoss}. The agent selects each prompt on a development set scored by sensitivity and specificity, the two quantities that determine how a note-level signal behaves once it is applied across a population. It consumes those development labels only to score and select prompts, never to update model weights, and it works from the same labeled budget that the lexicon threshold and the supervised comparator receive. This design targets the space between rule-based matching, which needs no labels but reads context poorly, and supervised extraction, which reads context but needs extensive annotation.

In this study we characterize the agent against two comparators across 72 signs and symptoms drawn from clinical notes: a curated rule-based lexicon, and a supervised BERT classifier fine-tuned per concept on the same 300-note development budget. To our knowledge this is the first evaluation of a loop-engineered extractor against a clinical detection task, with a rule-based and a supervised comparator, and it identifies the domain constraints that such a loop requires before clinical use. The supervised comparator demonstrates the annotation cost internally, rather than resting the claim on an external benchmark \cite{hu2026ie}. We report sensitivity and specificity rather than a single summary score, we group concepts by the operating-point shift the agent makes relative to the lexicon, and we measure how the prompt selected on the development set transfers to held-out validation. We also examine the agent's optimization behavior, including when it stops refining and retains its initial prompt. The analysis identifies where autonomous optimization improves extraction, where a lexicon remains preferable, and how the agent's operating point can be set to a clinical use.

\section{Background and related work}

Clinical information extraction converts narrative notes into structured variables, and rule-based systems defined the first generation of this work. Lexicon and pattern methods match curated term sets against the text, and pipelines such as cTAKES and CLAMP made them practical at scale \cite{savova2010ctakes,soysal2018clamp}. Their weakness is context, since terms such as ``chills'' appear in phrases beyond just acknowledging a symptom exists, such as negated statements, so a matcher keyed on the surface form cannot separate a present finding from other uses. As a result, handling negation, attribution, and temporality has occupied clinical natural language processing for two decades \cite{chapman2001negex,wang2018review,kreimeyer2017}.

Supervised neural models reduced this problem by learning context from annotated examples. Transformer encoders such as clinical BERT variants reached strong performance on entity recognition and concept extraction \cite{alsentzer2019clinicalbert,si2019contextual,gu2021pubmedbert}. A recent multi-institution study then established instruction-tuned large language models as the current state of the art, surpassing encoders most clearly in low-resource and cross-institution settings \cite{hu2026ie,chen2025benchmark}. That study also measured the cost of the gain. Its models required a manually annotated corpus of 1588 notes and parameter-efficient fine-tuning, and they ran 5 to 28 times slower than the encoder baseline at higher memory and energy \cite{hu2026ie}. The authors recommended encoders when labeled data are plentiful, which locates the binding constraint in the cost of annotation rather than the model architecture.

Large language models can also extract information without fine-tuning, through prompting. Few-shot prompting and manual prompt engineering improve clinical entity recognition without weight updates \cite{agrawal2022fewshot,hu2024prompt}, and instruction tuning extends this to biomedical entities \cite{keloth2024instruction}. Large language models have also been used to accelerate the annotation that supervised methods need \cite{goel2023annotation}, and broad benchmarks have mapped the strengths and weaknesses of these models across biomedical tasks \cite{chen2025benchmark,hu2026ie}. These approaches still depend on prompts that a person writes and revises. Automatic prompt optimization removes the manual step by searching instruction space against a development metric \cite{shin2020autoprompt,zhou2022ape,yang2023opro,khattab2023dspy}, yet existing clinical applications neither optimize the prompt toward a clinician-chosen operating point nor report how the selected prompt transfers to held-out data. Downstream tooling for generative extraction has begun to appear \cite{hsu2025llmie}, and local deployment of open-weights models has been demonstrated for privacy-preserving tasks such as de-identification \cite{wiest2025local,gptoss}, which shows that institution-internal extraction is feasible without sending notes to an external service.

Pythia occupies the space these methods leave open. It optimizes an extraction prompt for each concept autonomously, with no manual prompt engineering and no weight fine-tuning, and it selects the prompt on a development set scored by sensitivity and specificity \cite{estiri_pythia}. The model runs locally on open weights, so notes stay inside the institution \cite{gptoss}. The method removes the annotation and fine-tuning burden that supervised extraction requires \cite{hu2026ie,keloth2024instruction}, while supplying the context handling that rule-based matching lacks \cite{chapman2001negex,savova2010ctakes}. We introduced the agent and its optimization procedure previously \cite{estiri_pythia}. Here we characterize its behavior against a lexicon across 72 signs and symptoms, and we analyze where autonomous optimization helps, where it does not, and how its operating point can be set.

\section{Methods}
\subsection{Data and reference standard}

We evaluated the system on 400 clinical notes drawn from electronic health records. Clinicians annotated every note with binary labels for the presence or absence of each target sign or symptom, and these labels served as the reference standard. The set contained 72 signs and symptoms, with prevalence from below 2\% to above 20\%. Curated lexicons were made to cover these concepts, which define the head-to-head comparison reported here. For each concept we partitioned the 400 notes into a development set ($n=300$) and a held-out validation set ($n=100$) by stratified random sampling on binary symptom status, which preserved prevalence across the split. Partitioning was independent per concept, so a patient could fall in development for one concept and in validation for another.

The patient population of the study as seen in table \ref{tab:Population}, was selected randomly from a larger population of COVID patients. 400 notes were retrieved from the 387 patients and annotated by clinicians.

\begin{table}
    \centering
    \caption{Population statistics among the patient cohort used to test Pythia.}
    \label{tab:Population}
    \begin{tabular}{c|c}
    \toprule
         Patients & 387 \\
         Notes & 400 \\
         Mean Age (SD) & 58.8 (17.5) \\
         Median Age [Min, Max] & 60.0 [20.0, 95.0] \\
         Sex \\
         Female & 211 (54.5\%) \\
         Male & 176 (45.5\%) \\
         Race \\
         Asian & 11 (2.8\%) \\
         Black & 32 (8.3\%) \\
         Other & 53 (13.7\%) \\
         White & 282 (72.9\%) \\
         Ethnicity \\
         Hispanic & 25 (6.5\%) \\
         Non-Hispanic & 327 (84.5\%) \\
         Unknown & (9.0\%) \\
         \bottomrule
    \end{tabular}
\end{table}

\subsection{Symptom Selection}
The signs and symptoms annotated with binary labels were chosen because they are frequently present when patients are suffering from Long Covid, and the symptoms present in the set were determined by two clinicians.

\subsection{Autonomous prompt optimization}

\begin{figure}[t]
\centering
\includegraphics[width=\linewidth]{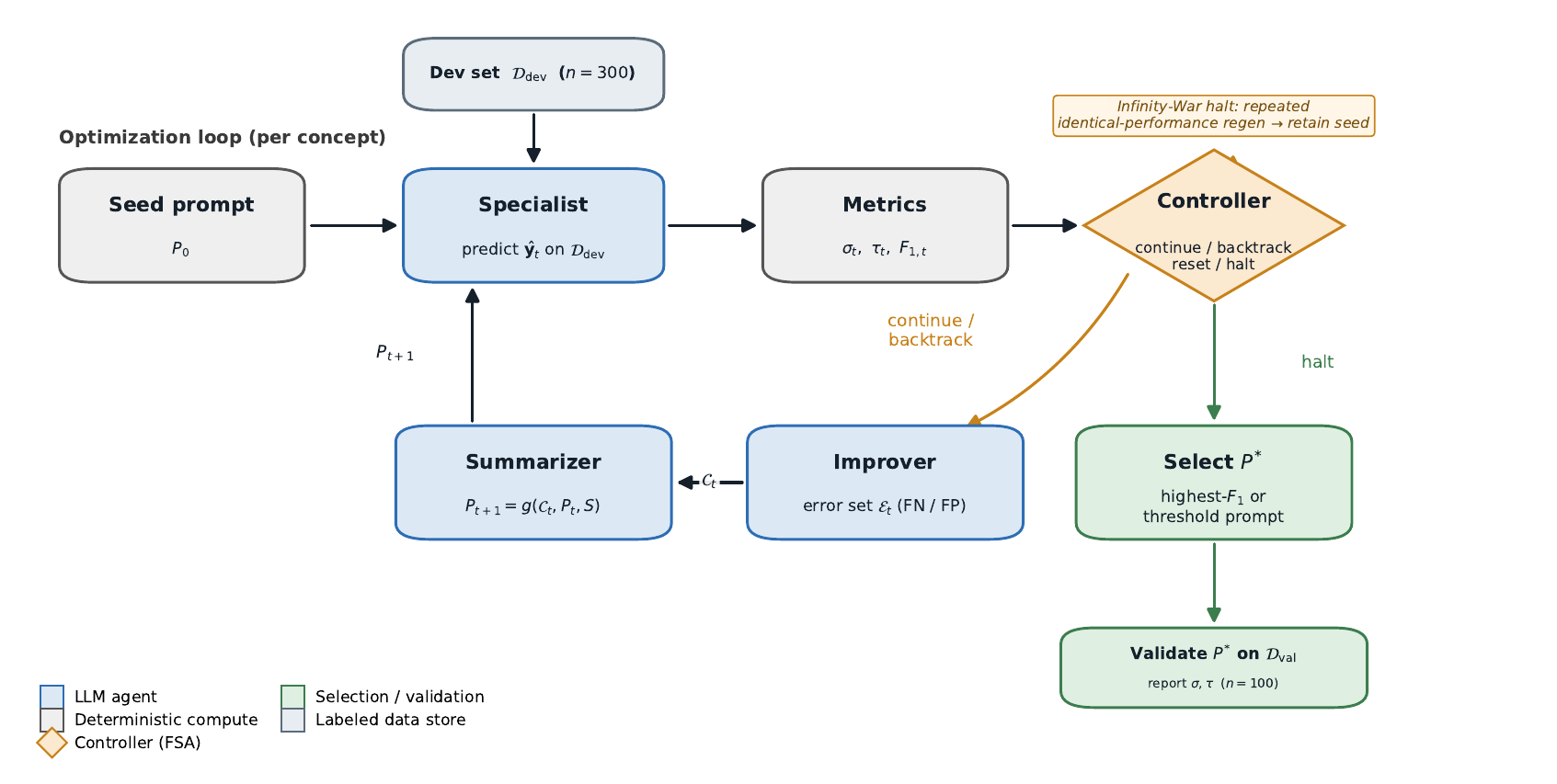}
\caption{The Pythia optimization loop. For each concept, a Specialist agent applies the current prompt $P_t$ to the development set and the loop computes sensitivity, specificity, and $F1$. A deterministic controller continues, backtracks to the best prior prompt, resets, or halts. On a non-halting step an Improver agent reads the errors for the active target, false negatives for a sensitivity target and false positives for a specificity target, and a Summarizer agent synthesizes the next prompt $P_{t+1}=g(\mathcal{C}_t,P_t,S)$. Repeated regeneration that recovers identical performance reaches the rejection cap and forces an Infinity~War halt. At halting the agent selects $P^{*}$ and applies it once to the held-out validation set.}
\label{fig:pipeline}
\end{figure}

Pythia searches the space of natural-language prompts for a per-concept note classifier (Figure~\ref{fig:pipeline}). Let $\mathcal{P}$ denote the space of prompt strings, $\mathcal{X}$ a corpus of notes, and $\mathcal{Y}=\{0,1\}$ the label for presence of a concept. Given a development set $\mathcal{D}_{\mathrm{dev}}\subset\mathcal{X}\times\mathcal{Y}$, the agent seeks a prompt
\begin{equation}
P^{*}=\operatorname*{arg\,max}_{P\in\mathcal{P}}\ \mathcal{M}\!\left(P;\mathcal{D}_{\mathrm{dev}}\right),
\end{equation}
where the score $\mathcal{M}$ is built from the confusion counts that prompt $P$ induces on $\mathcal{D}_{\mathrm{dev}}$. Writing $\mathrm{TP},\mathrm{FP},\mathrm{TN},\mathrm{FN}$ for those counts, sensitivity and specificity are $\sigma=\mathrm{TP}/(\mathrm{TP}+\mathrm{FN})$ and $\tau=\mathrm{TN}/(\mathrm{TN}+\mathrm{FP})$. We combine the two through a weighted operating-point objective
\begin{equation}
\mathcal{M}(P;\mathcal{D},w)=w\,\sigma(P;\mathcal{D})+(1-w)\,\tau(P;\mathcal{D}),\qquad w\in[0,1],
\end{equation}
where the weight $w$ sets the target trade-off, with $w\!\to\!1$ favoring case capture and $w\!\to\!0$ favoring false-positive control.
The head-to-head comparison reported here used a balanced objective, while the goal-setting analysis would sweep $w$ to trace each concept's attainable frontier.  The model is an open-weights large language model held fixed throughout, so no weights change and the prompt is the only object optimized, which places the method among automatic prompt-optimization procedures that search instruction space against a development metric \cite{shin2020autoprompt,zhou2022ape,yang2023opro,khattab2023dspy}.

Pythia is a multi-agent loop implemented as a stateful directed graph $\mathcal{G}=(V,E)$ with cycles \cite{langgraph}, in which each node is a function $f_i:\mathcal{S}\to\Delta\mathcal{S}$ that reads a shared state $\mathcal{S}$ and returns a partial update \cite{pydantic}. A Specialist agent applies the current prompt $P_t$ to every note in $\mathcal{D}_{\mathrm{dev}}$ and returns predictions $\hat{\mathbf{y}}_t\in\{0,1\}^{n}$, from which the loop computes $\sigma_t$, $\tau_t$, $\mathrm{accuracy}_t$ and $F1_t$. An Improver agent then reads the misclassified notes for the active target. For a sensitivity target the error set is the false negatives, $\mathcal{E}_t=\{x_i:\hat{y}_{t,i}=0\wedge y_i=1\}$; for a specificity target it is the false positives, $\mathcal{E}_t=\{x_i:\hat{y}_{t,i}=1\wedge y_i=0\}$. The Improver produces a natural-language critique of each error, and a Summarizer agent synthesizes the next prompt,
\begin{equation}
P_{t+1}=g\!\left(\mathcal{C}_t,\,P_t,\,S\right),
\end{equation}
where $\mathcal{C}_t$ is the critique set and $S$ a fixed operating procedure. When a step lowers development $F1$, the agent passes the regressing prompt to the Summarizer as a negative example, so synthesis avoids the change that caused the regression.

In loop-engineering terms, the Specialist, Improver, and Summarizer are the sub-agents that propose a prompt, the controller is the separate process that checks the stopping condition so the agent proposing a prompt does not judge its own halt, and the iteration and transition histories are the external memory that persists state across iterations.

\subsection{Controller and termination}

A deterministic controller reads the trajectory history and returns one of four decisions, continue, backtrack, reset, or halt (Algorithm~\ref{alg:controller}). The controller halts when both targets clear preset thresholds, $\sigma_t\ge\theta_\sigma$ and $\tau_t\ge\theta_\tau$, when the iteration count reaches $T_{\max}$, or when consecutive rejected refinements reach a cap $k_{\max}$. It backtracks to the highest-$F1$ prompt when a refinement regresses development $F1$ by more than a tolerance $\epsilon$. Repeated regeneration from an early iteration that recovers identical performance reaches the rejection cap and forces termination, the behavior we term an Infinity~War halt. We set $\epsilon=0.05$ and $k_{\max}=2$.

\begin{algorithm}[t]
\caption{Controller decision at iteration $t$}
\label{alg:controller}
\begin{algorithmic}[1]
\Require metrics $\sigma_t,\tau_t,F1_t$; history $H$; iteration $t$; thresholds $\theta_\sigma,\theta_\tau$; limits $T_{\max},k_{\max}$; rejection count $k$; tolerance $\epsilon$
\If{$\sigma_t \ge \theta_\sigma$ \textbf{and} $\tau_t \ge \theta_\tau$}
    \State \Return $\langle\, \text{halt},\ P^{*}\leftarrow P_t \,\rangle$
\ElsIf{$t \ge T_{\max}$ \textbf{or} $k \ge k_{\max}$}
    \State \Return $\langle\, \text{halt},\ P^{*}\leftarrow \arg\max_{i} F1_i \,\rangle$
\ElsIf{$|H|\ge 2$ \textbf{and} $F1_t < (1-\epsilon)\,F1_{t-1}$}
    \State $t^{*}\leftarrow \arg\max_{i} F1_i$
    \State \Return $\langle\, \text{backtrack},\ P_t\leftarrow P_{t^{*}},\ k\leftarrow k+1 \,\rangle$
\Else
    \State \Return $\langle\, \text{continue},\ k\leftarrow 0 \,\rangle$
\EndIf
\end{algorithmic}
\end{algorithm}

\subsection{Prompt selection and evaluation}

When the controller halts on the threshold condition, the active prompt becomes $P^{*}$. Under any other halt the agent selects the prompt with the highest development $F1$ across the trajectory. We then applied $P^{*}$ once to the held-out validation set and report validation sensitivity and specificity. We optimized and report sensitivity and specificity rather than a single summary score, because the two methods occupy different regions of the operating-point plane, and a scalar such as $F1$ would obscure that distinction. $F1$ is used only as the internal tie-break for selection and backtracking, since the F1 score represents the best overall performing prompt in the development dataset to build off of and optimize towards the specific metrics a user requires.

\subsection{Model selection}

We selected the language model used for the complete set of signs and symptoms via a preliminary phase that compared three open-weights models: Gemma-4-31B-it \cite{gemma4}, GPT-OSS-20B \cite{gptoss}, and Llama-3.1-70B \cite{llama31} on five signs and symptoms spanning a range of prevalence (Table~\ref{tab:modelselect}). GPT-OSS-20B was selected for the full evaluation because it was the only model to reach perfect validation sensitivity across all five concepts (mean 1.00, versus 0.97 for Gemma~4 31B and 0.66 for Llama~3.1 70B), and it produced the highest mean development F1 (0.67), the metric the controller itself uses for judging the right backtracking prompt. Gemma~4 31B matched GPT-OSS-20B on mean validation F1 but only by trading sensitivity for specificity on individual concepts, when false negatives are the costlier error type for a symptom detector. All models were run locally through an inference server at temperature 0 for deterministic output, and each optimization trajectory was initialized from the seed prompt \textit{``Is the patient showing signs of [symptom]?''} with no additional clinical detail supplied.

\subsection{Lexicon}
For each of the 72 symptom categories, we constructed a rule-based lexicon of category-specific terms and applied it to note-level text using exact substring matching with word boundaries. Candidate matches were filtered for negation and family-history context: matches preceded within a 50-character window by a negation cue (e.g., "no," "denies," "without," "resolved") or a family-history reference (e.g., "mother," "family history") were excluded from the symptom count, consistent with standard approaches to discounting negated or non-patient-referent mentions in clinical text. For each condition, a per-note mention count was thresholded to produce a binary prediction, with the threshold selected to maximize F1 score on the same dataset. Performance was evaluated separately on the development and validation sets for each condition, with sensitivity, specificity, accuracy, and F1 score. Ninety-five percent confidence intervals for all metrics were derived from 1,000 stratified bootstrap resamples, in which positive and negative notes were resampled independently to preserve class prevalence.

\subsection{Supervised comparator (BERT)}
\label{sec:bert}

For each symptom, we fine-tuned a binary classifier per-condition with Bio\_ClinicalBERT \cite{alsentzer2019clinicalbert}, a BERT based encoder trained on MIMIC-III clinical notes. Because clinical notes frequently exceed BioClinicalBERT's 512-token context limit, each note was segmented into overlapping 512-token windows (stride 64 tokens), with chunk-level labels assigned from character-level span annotations so that only chunks containing a confirmed symptom mention were labeled positive. Models were trained using class-weighted cross-entropy loss and stratified 5-fold cross-validation on 300 development-set notes per condition, with early stopping (patience 2 epochs) selecting the best checkpoint per fold. Out-of-fold predictions were pooled across folds as the primary cross-validated performance estimate. The final model was then retrained on all 300 development-set notes using the same configuration and evaluated on the validation dataset never used during training or model selection. Training used a learning rate of $ 2 \times 10^{-5}$, batch size 8, weight decay 0.01, and a maximum of 5 epochs. Sensitivity, specificity, F1 score, and accuracy with 95\% stratified bootstrap confidence intervals (1,000 resamples) were reported for the cross-validated development and validation datasets.

\subsection{Comparator and Analysis}
The curated lexicon and the BERT classifier were used as comparators against Pythia performance. We computed sensitivity and specificity from validation confusion counts and report 95\% confidence intervals via bootstrap. For each concept $c$ we summarized the contrast against the lexicon as the operating-point shift $\Delta_c=(\Delta\sigma_c,\Delta\tau_c)=(\sigma^{\mathrm{Pythia}}_c-\sigma^{\mathrm{lex}}_c,\ \tau^{\mathrm{Pythia}}_c-\tau^{\mathrm{lex}}_c)$, and we grouped concepts by the sign and magnitude of $\Delta_c$ into the four regions reported below. We measured development-to-validation transfer as the per-metric gap $\delta=r_{\mathrm{dev}}-r_{\mathrm{val}}$ for $r\in\{\sigma,\tau\}$ and related $\delta$ to concept prevalence.

\section{Results}

\subsection{Overall detection performance}

\begin{figure}[t]
\centering
\includegraphics[width=0.82\linewidth]{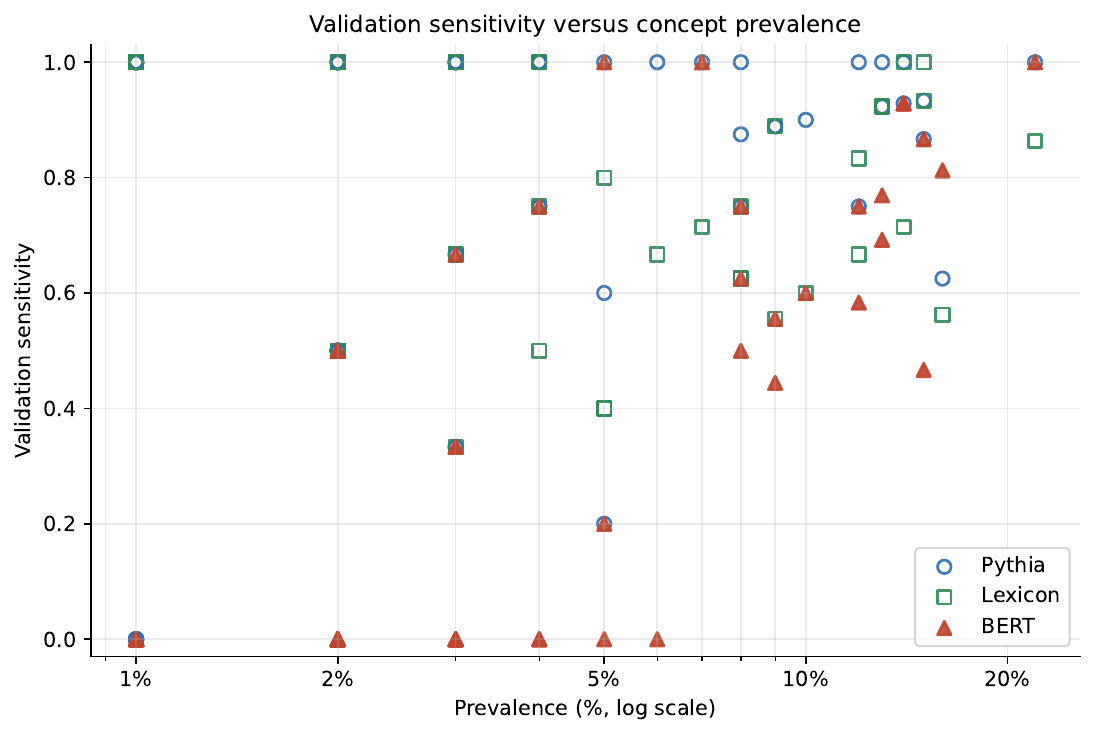}
\caption{Validation sensitivity versus concept prevalence (log scale) for Pythia, the lexicon, and BERT, across all 72 concepts. BERT sensitivity clusters near zero below roughly 5\% prevalence, while Pythia and Lexicon points remain mostly above 0.5 across the full range.}
\label{fig:bertprev}
\end{figure}

\begin{figure}[t]
\centering
\includegraphics[width=0.82\linewidth]{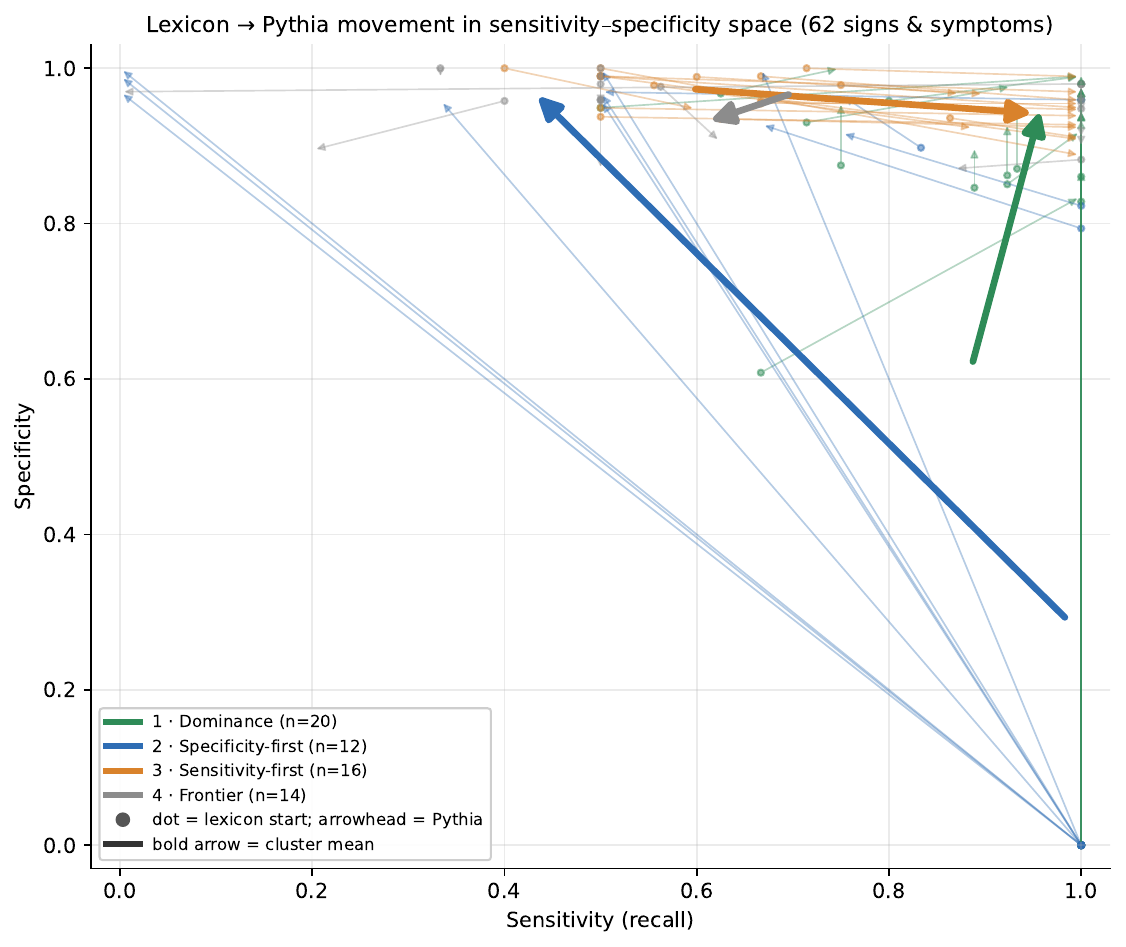}
\caption{Displacement of each sign or symptom in the sensitivity--specificity plane from the lexicon (open marker) to Pythia (arrowhead), for the 62 concepts. Color encodes the four groups of Table~\ref{tab:clusters}; bold arrows give group means. The cluster of lexicon markers along the lower-right edge, at sensitivity near 1.0 and specificity near 0, corresponds to concepts on which the lexicon labels every note positive; the specificity-first arrows lift these into a usable operating region.}
\label{fig:ssplane}
\end{figure}

We evaluated Pythia against a curated lexicon on 72 signs and symptoms, scoring each method once on a held-out validation set after prompt selection on a separate development set. Across all 72 concepts, Pythia reached a mean sensitivity of 0.76 and a mean specificity of 0.95, whereas the lexicon reached a mean sensitivity of 0.82 and a mean specificity of 0.76, and the fine-tuned BERT classifier achieved a mean sensitivity of 0.23 and a mean specificity of 0.91. Mean accuracy for BERT was 0.90, comparable to Pythia's and the lexicon's, despite this sensitivity deficit. The high accuracy was a consequence of the negative class dominating most concepts and illustrates why we report sensitivity and specificity separately rather than a single summary score (Section 3.5). BERT's sensitivity was exactly zero on 47 of 72 concepts, a failure mode distinct from the lexicon's. The lexicon is overly sensitive, as on 14 concepts the lexicon labels every note as positive (Section 4.2), BERT under-reports, missing every positive case on concepts it had too few positive examples to learn from. \ref{fig:bertprev} Thirteen of the 14 concepts where the lexicon fails are among the 47 where BERT fails, the same low-prevalence concepts defeating both comparators through opposite mechanisms. Because Pythia optimizes sensitivity and specificity directly, we characterized its behavior as displacement in the sensitivity-specificity plane relative to the lexicon (Figure~\ref{fig:ssplane}).

\subsection{A sensitivity-specificity taxonomy of autonomous optimization}
\label{sec:taxonomy}

\begin{table}[t]
\centering
\caption{Validation performance of Pythia, the lexicon, and a fine-tuned BERT comparator across the four sensitivity-specificity groups (62 signs and symptoms). Values are group means. $\Delta$ denotes Pythia minus lexicon.}
\label{tab:clusters}
\begin{tabular}{lccccccccc}
\toprule
& & \multicolumn{2}{c}{Pythia} & \multicolumn{2}{c}{Lexicon} & \multicolumn{2}{c}{$\Delta$ (Pythia$-$lexicon)} & \multicolumn{2}{c}{BERT} \\
\cmidrule(lr){3-4} \cmidrule(lr){5-6} \cmidrule(lr){7-8} \cmidrule(lr){9-10}
Group & $n$ & Sens. & Spec. & Sens. & Spec. & $\Delta$Sens. & $\Delta$Spec. & Sens. & Spec. \\
\midrule
Dominance & 20 & 0.96 & 0.95 & 0.89 & 0.62 & 0.07 & 0.33 & 0.31 & 0.89 \\
Specificity-first & 12 & 0.43 & 0.97 & 0.99 & 0.29 & -0.56 & 0.68 & 0.13 & 0.96 \\
Sensitivity-first & 16 & 0.95 & 0.94 & 0.59 & 0.97 & 0.36 & -0.03 & 0.32 & 0.85 \\
Frontier & 14 & 0.61 & 0.93 & 0.7 & 0.97 & -0.09 & -0.04 & 0.25 & 0.91 \\
\bottomrule
\end{tabular}
\end{table}

Setting aside ten unambiguous single-token findings for which lexical matching remained preferable (Section~\ref{sec:lexwins}), the remaining 62 concepts partitioned into four groups defined by the sign of the change in each metric (Table~\ref{tab:clusters}). The partition is mechanistic rather than performance-binned, because each group corresponds to a distinct operating-point move that the agent made under a single objective.
 
In the largest group (\(n=20\)), Pythia raised both sensitivity and specificity above the lexicon, by 0.07 and 0.33 on average. These concepts spanned common presentations such as abdominal pain, chest pain, cough, and weakness, together with context-dependent findings such as homicidal ideation and urinary frequency. For abdominal pain, sensitivity rose from 0.71 to 0.93 and specificity from 0.93 to 0.98; for paresthesia, sensitivity rose from 0.80 to 1.00 at a specificity of 0.99. Across this group autonomous optimization improved detection on both axes at once, without the loss of sensitivity that ordinarily accompanies a specificity gain.
 
\begin{figure}[t]
\centering
\includegraphics[width=0.85\linewidth]{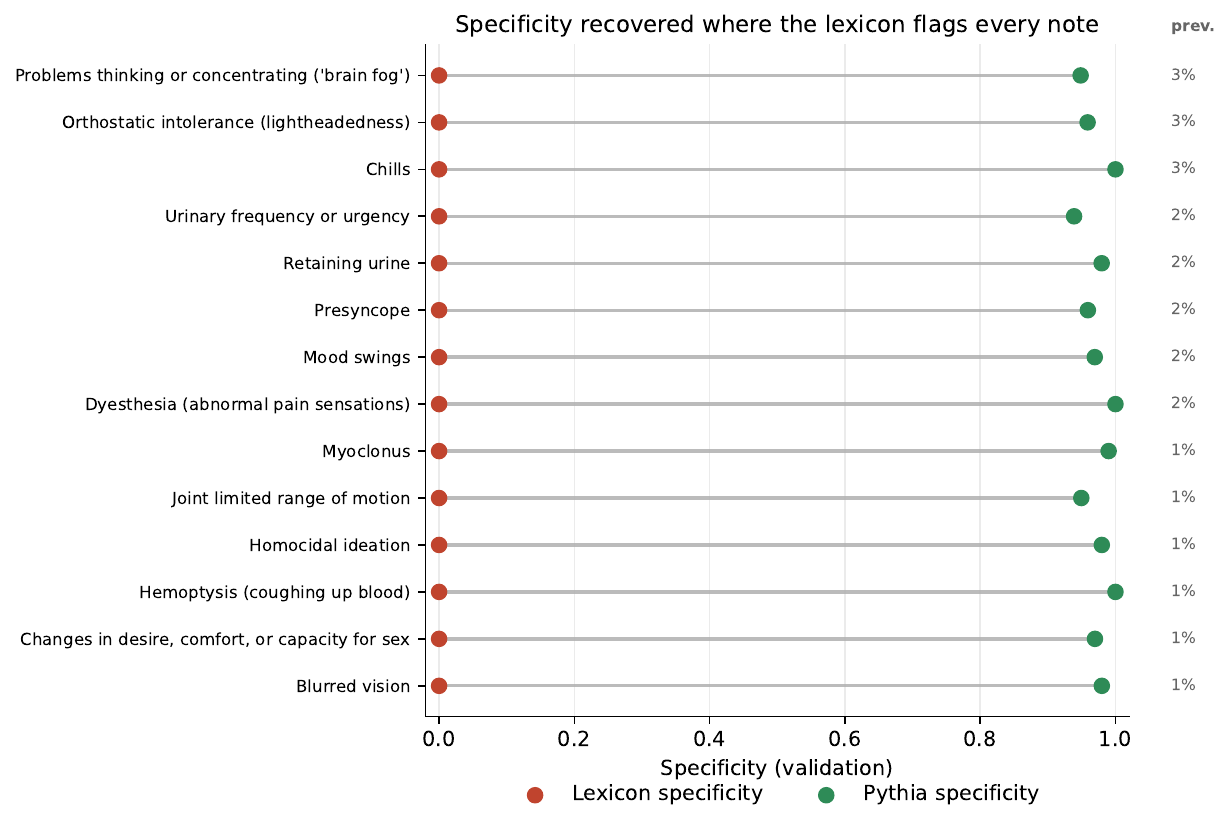}
\caption{Specificity recovered on the 14 concepts where the lexicon returns a specificity of zero, ordered by prevalence. Each row connects the lexicon specificity at zero to the Pythia specificity on the same held-out notes, with prevalence annotated at the right. For these concepts the matched term appears in negated, hypothetical, templated, or historical context, so lexical matching labels every note positive; Pythia raised specificity to a mean of 0.97 by conditioning detection on a present-tense patient finding.}
\label{fig:speclift}
\end{figure}

A second group (\(n=12\)) showed the agent suppressing false positives at the cost of recall. Here the lexicon fired on nearly every note, with a mean specificity of 0.29 against a mean sensitivity of 0.99, whereas Pythia raised specificity to 0.97 and reduced sensitivity to 0.43. The recall that Pythia surrendered was largely artifactual, because the lexicon reached near-perfect sensitivity only by labeling all notes positive. Chills illustrates the shift, moving from a lexicon specificity of 0.00 to a Pythia specificity of 1.00 as the agent revised the prompt to require a present-tense patient finding rather than any textual mention of the term. Across the 14 concepts on which the lexicon returned a specificity of zero, which span this group and the dominance group, Pythia raised specificity to a mean of 0.97 (Figure~\ref{fig:speclift}).
 
A third group (\(n=16\)) showed the converse adjustment, with Pythia recovering missed cases while holding specificity nearly constant. Sensitivity rose by 0.36 on average against a specificity change of \(-0.03\). For headache, sensitivity rose from 0.62 to 1.00 at a specificity near 0.92, and for confusion from 0.71 to 1.00 at a specificity of 0.99. The agent extended detection to phrasings absent from the fixed vocabulary at minimal false-positive cost, the move that most closely approximates Pareto improvement among the trade-off groups.
 
In the remaining group (\(n=14\)), Pythia neither dominated the lexicon nor cleanly exchanged one metric for the other, with mean changes of \(-0.09\) in sensitivity and \(-0.04\) in specificity. These concepts were sparse and semantically diffuse, including malaise, suicidal ideation, and trauma-related symptoms, and they delineate the boundary of what unsupervised optimization recovered from the development set. We treat this group as the frontier of the current method and return to its implications in the Discussion.

BERT achieved a 0.31 mean sensitivity and 0.89 mean specificity on the Pythia-dominated group, with 0.13 mean sensitivity and 0.96 mean specificity on the 12-symptom group, where the agent suppressed false positives at a cost of recall, and a mean sensitivity of 0.32 and 0.85 mean specificity on symptoms where Pythia had found missed cases over maintaining true negatives. Finally, on the remaining group where the lexicon outperformed Pythia, it achieved a mean sensitivity of 0.25 and a mean specificity of 0.91 on the frontier group. BERT's sensitivity is low and roughly flat across all four groups, ranging between 0.13 and 0.32, so BERT's comparatively low performance is not tied to any characteristic of the system's operating point, unlike Pythia and the lexicon.

\subsection{Concepts where lexical matching remained preferable}
\label{sec:lexwins}

For ten findings named by a single unambiguous token, including tremor, melena, and hematuria, exact lexical matching reached a mean specificity of 0.98 and a mean sensitivity of 0.95, and Pythia did not improve on it, attaining a mean sensitivity of 0.60. These concepts carried a mean prevalence of 0.01, and their surface terms map almost one-to-one onto the finding, so string matching is both sensitive and specific and leaves no contextual ambiguity for the agent to resolve. BERT fared worse than either comparator here, with a mean sensitivity of exactly zero across all ten concepts, consistent with the low positive-example counts available at this prevalence (Section~\ref{sec:prevfloor}). We analyze this group separately because it defines the regime in which an autonomous agent confers no advantage over a lexicon.

\subsection{Development-to-validation transfer}
\label{sec:transfer}

\begin{figure}[t]
\centering
\includegraphics[width=0.95\linewidth]{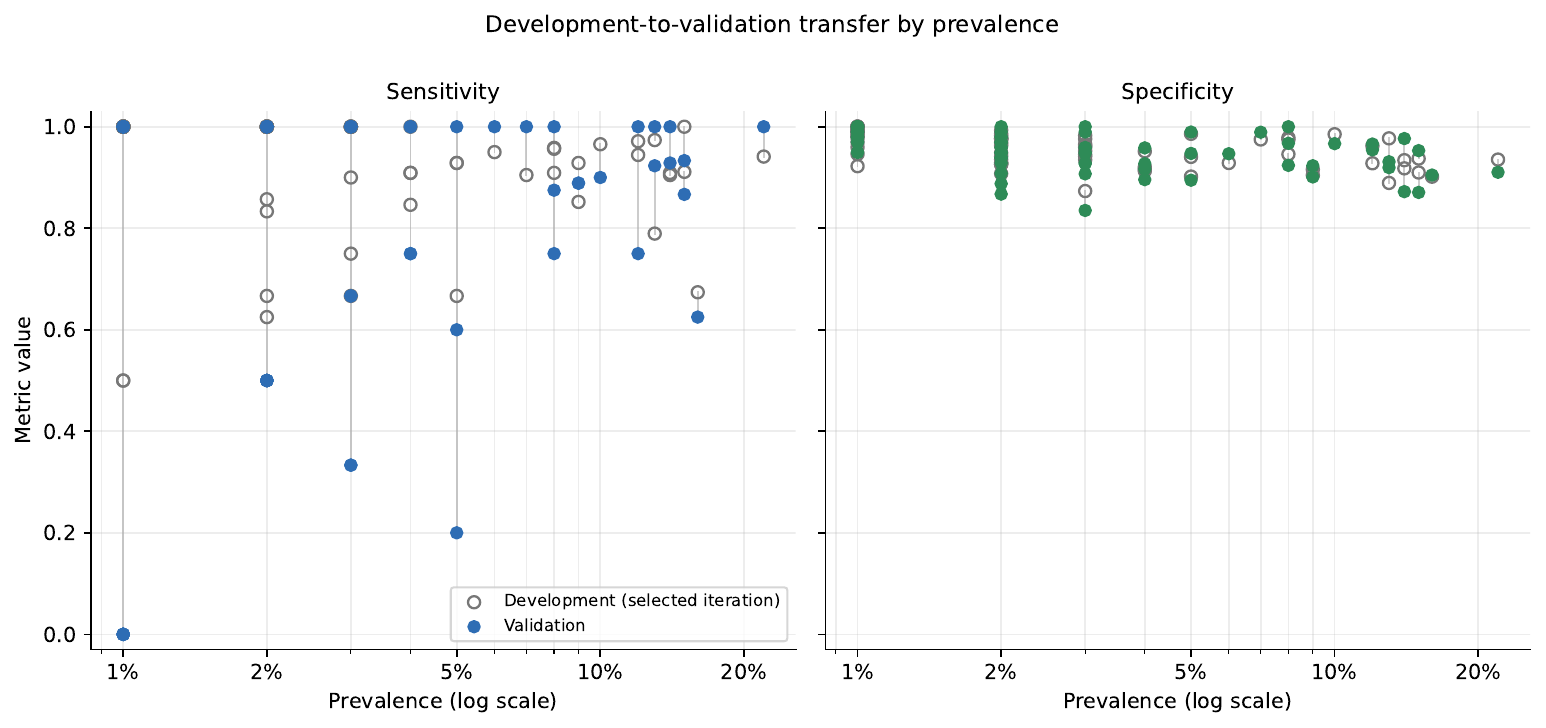}
\caption{Development-to-validation transfer by prevalence. For each of the 72 concepts, the open marker gives the selected development iteration and the filled marker gives validation, plotted against prevalence on a logarithmic axis. Specificity (right) transfers with a near-zero gap across the full prevalence range, because it is estimated against a large negative class. Sensitivity (left) transfers above 5\% prevalence but is overstated on the development set as prevalence falls, reaching a mean gap of 0.25 below 2\% prevalence, where positive examples are sparse.}
\label{fig:transfer}
\end{figure}

We then asked whether the operating point selected on the development set transferred to held-out validation. Specificity transferred almost exactly, with a mean development-to-validation gap of 0.004 that did not vary across prevalence strata. Sensitivity transferred well for concepts above 5\% prevalence, where the gap was 0.015 or smaller, yet degraded as prevalence fell, reaching 0.25 below 2\% prevalence. Because the development set contained only a few positive examples for the rarest concepts, the sensitivity estimated during optimization overstated held-out sensitivity, whereas specificity, estimated against a large negative class, remained stable. The transfer behavior therefore tracks the denominator of each metric rather than any property of the agent, and it identifies a prevalence floor below which development-set selection of sensitivity is unreliable (Figure~\ref{fig:transfer}).

\subsection{A prevalence floor for supervised fine-tuning}
\label{sec:prevfloor}
 
BERT's sensitivity tracked concept prevalence far more sharply than either Pythia's or the lexicon's (Spearman $\rho=0.75$ with prevalence; $\rho=0.75$ with the number of positive development notes). Below 2\% prevalence (37 concepts, a mean of 4.6 positive development notes each), BERT's mean sensitivity was 0.03, and sensitivity was exactly zero for 95\% of these concepts. Mean sensitivity rose to 0.19 at 2--5\% prevalence (17 concepts, mean 10.5 positive notes), 0.56 at 5--10\% (8 concepts, mean 24.3 positive notes), and 0.78 above 10\% (10 concepts, mean 43.7 positive notes), where no concept showed zero sensitivity (Table~\ref{tab:prevfloor}, Figure~\ref{fig:bertprev}).
 
This pattern differs mechanistically from Pythia's prevalence-dependent transfer gap (Section~\ref{sec:transfer}). Pythia's development-set sensitivity is optimistic at low prevalence and overstates held-out performance; BERT's cross-validated development-set sensitivity pooled from predictions across five folds (Section~\ref{sec:bert}) is itself near zero at low prevalence (mean 0.02 below 5\% prevalence), indicating the classifier fails to learn a usable signal even within cross-validated development, rather than merely failing to transfer a dev-fit signal to held-out validation. Five-fold splitting compounds the scarcity: a concept with three to five positive notes across the full 300-note development set may leave individual folds' training portions with only one or two positive examples, little signal for gradient-based fine-tuning to shift the decision boundary away from the majority class.
 
The two comparators fail in opposite directions at low prevalence. The lexicon is overly sensitive, returning a specificity of zero on 14 concepts because a matched term appears in negated, hypothetical, or templated contexts (Section~\ref{sec:taxonomy}, Figure~\ref{fig:speclift}); BERT under-reports, returning a sensitivity of zero on 47 concepts because too few positive examples exist to fine-tune on. Pythia is not immune to low prevalence, as its own sensitivity transfer degrades below 5\% prevalence (Section~\ref{sec:transfer}). Even with reduced performance, Pythia does not collapse to zero on any concept, consistent with prompt optimization requiring no labeled examples beyond what the development set provides for scoring.

\subsection{Optimization dynamics}

Pythia refined each prompt over a median of two development iterations, reaching eight at most. On 17 of 72 concepts the controller backtracked to an earlier iteration when refinement degraded development performance. On 8 concepts it reached the Infinity~War termination, in which the agent regenerated a prompt from an early iteration and recovered the same measured performance, then repeated the attempt without change and advanced to the next concept at the iteration limit. The repeated, performance-invariant regenerations logged for these eight concepts mark the development items on which continued optimization conferred no measurable gain, and they coincide with the frontier group rather than with the concepts on which Pythia improved over the lexicon. Figure~\ref{fig:traj} contrasts the two outcomes at matched prevalence, with refinement recovering sensitivity at held specificity for one concept and failing to improve on the seed for another, after which the agent retained the seed.

\section{Discussion}

An autonomous agent that wrote and optimized its own extraction prompts matched a curated lexicon on case capture while raising specificity across 72 signs and symptoms. It separated from two comparators that failed in opposite directions, the lexicon by over-firing on the negative class and the supervised classifier by under-firing at low prevalence. Pythia reached a mean sensitivity of 0.76 against the lexicon's 0.82, and a mean specificity of 0.95 against 0.76, and it equaled or exceeded the lexicon on both metrics for 20 concepts. A single summary score such as $F1$ hides this pattern, because the two methods occupy different regions of the sensitivity--specificity plane and the decisions that follow from note-level extraction depend on the two metrics separately. At the prevalence of most symptoms a specificity gain of this size decides whether an extracted signal is usable, since a method that fires on every note returns mostly false positives once the negative class is large.

Our earlier report characterized the optimization procedure and its instability in the general agentic setting \cite{estiri_pythia}. Here we evaluate the same system as a clinical extractor, contributing the sensitivity-specificity taxonomy, the transfer analysis from development to validation, and the comparison against both a lexicon and a supervised classifier.

The fine-tuned supervised comparator sharpens this picture further. This comparator reflects a deliberately small per-concept fine-tuning budget matched to the labels that Pythia and the lexicon receive, and it therefore characterizes supervised extraction under scarce annotation rather than the performance reported with thousands of annotated notes. Under the same 300-note development budget used throughout this study, BERT reached a mean sensitivity of 0.23 against Pythia's 0.76 and the lexicon's 0.82, and collapsed to zero sensitivity on concepts below roughly 5\% prevalence (Section~\ref{sec:prevfloor}). This gives the annotation-burden claim in the introduction a demonstration internal to this study, rather than resting solely on an external benchmark that required 1,588 annotated notes to fine-tune competitively \cite{hu2026ie}. It also illustrates, concretely, why we report sensitivity and specificity rather than accuracy: BERT's mean accuracy (0.90) looked competitive with Pythia's and the lexicon's despite detecting zero true positives on 47 of 72 concepts, because accuracy is dominated by the negative class at low prevalence.

The specificity gain reflects the agent recovering context handling that clinical extraction has historically obtained through annotation or hand-built rules \cite{chapman2001negex,savova2010ctakes,si2019contextual}. On the 14 concepts where the lexicon labeled every note positive, Pythia raised specificity to a mean of 0.97 by rewriting the prompt to require a present-tense, patient-attributed finding rather than any textual mention of the term (Figure~\ref{fig:speclift}). The supervised benchmark that defines the current state of the art encodes this same information as labeled modifiers for negation, temporality, and subject \cite{hu2026ie,uzuner2011i2b2}. Pythia reconstructs those distinctions for each concept without labels for them, which is the property that lets it extend to many concepts without a matching annotation effort.

The agent did not apply one fixed trade-off. Across the 62 directly compared concepts, it occupied four regions of the sensitivity-specificity plane (Figure~\ref{fig:ssplane}; Table~\ref{tab:clusters}). It dominated the lexicon on both metrics for 20 concepts, raised specificity at the cost of recall for 12, raised sensitivity at near-constant specificity for 16, and fell behind for 14. Because the optimization objective is an input to the agent rather than a fixed property, the spread across these four regions shows that one objective already places different concepts at different operating points. Whether a clinician can set that operating point deliberately, by weighting the development objective toward sensitivity or toward specificity, follows as a direct prediction of this design. We have not yet tested it; the weighted objective in Equation 2 defines the experiment, and a per-concept sweep of the weight is the immediate next step \cite{yang2023opro,zhou2022ape}. A lexicon offers no such control, and a supervised model requires retraining to move \cite{hu2026ie}.

Selection on the development set transferred to validation for specificity but not uniformly for sensitivity. The specificity gap between development and validation averaged 0.004 and did not vary with prevalence, whereas the sensitivity gap widened from minimal above 5\% prevalence to 0.25 below 2\% (Figure~\ref{fig:transfer}). The pattern follows the denominator of each metric. Specificity is estimated against a large negative class and stays stable, while sensitivity rests on a few positive examples at low prevalence and is overstated during optimization. This identifies a prevalence floor below which development-set sensitivity should not be trusted, and it argues for human review or larger development sampling for the rarest concepts. The present evaluation used one institution's notes, which limits external validity but also avoids the test-set contamination that affects public corpora likely present in model pretraining \cite{hu2026ie,chen2025benchmark}. We are extending the evaluation across sites in a federated network, where prompts are shared and scored locally without moving notes, which tests cross-institution transfer under the privacy constraints that motivate local deployment \cite{wiest2025local}.

Two groups of concepts mark the limits of the approach. For ten findings named by a single unambiguous token, such as tremor and melena, exact matching reached a mean specificity of 0.99 and a mean sensitivity of 0.95, and the agent did not improve on it and sometimes underperformed by over-specifying the prompt. For a frontier group of sparse, semantically diffuse concepts, refinement did not improve on the seed. These results support a division of labor at deployment: route unambiguous, high-specificity terms to a lexicon \cite{savova2010ctakes,soysal2018clamp}, and reserve the agent for concepts whose surface form is ambiguous, where context reading is the source of its advantage.

Pythia's optimization includes a behavior that matters for autonomous use, because it recognizes when refinement cannot improve and stops. On 8 concepts the controller regenerated a prompt from an early iteration, recovered the same development performance without change, and terminated under the Infinity~War criterion, retaining the seed rather than shipping a degraded prompt (Figure~\ref{fig:traj}). An autonomous loop that revises its own instructions can fail silently by drifting or looping, a known failure mode of prompt-optimization search \cite{yang2023opro,zhou2022ape}, and a detector that halts unproductive search guards against it. We report this behavior because it bears on whether such agents can run without supervision, beyond the task studied here.

This study set out to test whether a loop-engineered extractor can meet the sensitivity and specificity requirements that clinical use imposes. The results give a qualified answer. Across 62 directly compared concepts the loop maintained or improved specificity without additional annotation, recovered context distinctions that rule-based matching cannot supply, and transferred its selected operating point to held-out validation above a prevalence floor of 5 percent. The loop also detected when refinement could not improve and halted, the property that lets an unattended loop avoid silent degradation. Three domain constraints qualify the result and define what a clinical loop requires: a prevalence floor below which the development set is too sparse to trust the selected prompt, a fallback to exact matching for unambiguous single-token terms, and human review at the rare tail. These constraints locate the human where loop engineering places them, as the designer and verifier of the loop rather than the writer of each prompt.

Against rule-based extraction, the agent supplies context handling that a fixed term set cannot provide \cite{chapman2001negex,savova2010ctakes}. Against supervised extraction, it removes the annotation and fine-tuning that the current state of the art requires \cite{hu2026ie,keloth2024instruction}, while running locally on open weights \cite{wiest2025local,gptoss}. The cost moves rather than disappears, because optimization spends repeated inference passes during the per-concept search, so the saving is in annotation and training rather than in total computation, and we report the search budget so the trade is explicit. The practical consequence is that coverage can extend to many concepts without a proportional annotation effort, with the advantage concentrated where concepts are ambiguous and absent where they are unambiguous.

Several limitations qualify these findings. We evaluated one open-weights model at a single institution, so model and site generalization remain open. The cohort consists of COVID-19 patients and the concept set was chosen for post-acute COVID-19, so its prevalence distribution and documentation patterns may not represent other clinical populations. Behavior at the operating point may also depend on the backbone, because the model-selection phase showed Gemma 4 31B trading sensitivity for specificity on individual concepts where GPT-OSS-20B did not. The validation sets are small, and the intervals are wide under low prevalence. So, several concept-level differences rest on one or two cases. We optimized and reported sensitivity and specificity, while precision and net clinical benefit were not optimization targets, and precision at low prevalence is sensitive to a few false positives. The reference standard and its agreement set the ceiling on measurable performance. We adjudicated that standard with clinicians but did not compute a formal inter-annotator agreement, so the residual label noise is uncharacterized. Detection of a sign or symptom is a surrogate for any downstream use, which we have not yet demonstrated. The supervised comparator we report (Section~\ref{sec:bert}) reflects a small per-concept fine-tuning budget, not the state of the art; Hu et al.\ fine-tuned on 1,588 annotated notes and reported substantially higher performance \cite{hu2026ie}. Our BERT result characterizes supervised fine-tuning under the same annotation budget available to Pythia and the lexicon, not an upper bound on what supervised extraction can achieve with adequate data, and its performance should not be generalized to supervised extraction under adequate training data.

Four steps follow from this work. The objective sweep will establish whether the operating point is controllable rather than only variable. Federated evaluation across sites will test cross-institution transfer without moving notes. A supervised comparator and a larger-model ceiling will place the agent on the spectrum from rules to fine-tuned extraction \cite{hu2026ie,keloth2024instruction}. A downstream phenotyping task will test whether agent-extracted symptoms support a clinical endpoint, the evidence that detection accuracy alone cannot provide \cite{hripcsak2013,banda2018}. The study supports a measured claim, that autonomous, fine-tuning-free, locally run prompt optimization can build symptom extractors that favor specificity, generalize from development to validation outside the rare tail, and expose an operating point that can be set to the clinical use.

\section{Acknowledgments}

\paragraph*{Data and Code Availability}
This paper uses a dataset of 400 chart-reviewed EHR notes, each annotated with binary yes/no classifications on various signs and symptoms by clinicians. The dataset is confidential and not publicly available. The final metrics for each concept can be found within the supplementary data file. The code can be found \href{https://github.com/clai-group/Pythia}{here}.

\paragraph*{Ethics approval}
This study was approved by the Mass General Brigham Institutional Review Board (IRB) Protocol 2020P001063.

\paragraph*{Funding}
This study has been supported by grants from the National Institutes of Health: the
National Institute on Aging (RF1AG074372) and the National Institute of Allergy and Infectious
Diseases (R01AI165535)

\pagebreak

\section*{Supplementary Material}
\renewcommand{\thetable}{S\arabic{table}}
\renewcommand{\thefigure}{S\arabic{figure}}
\begin{table}[H]
\setcounter{table}{0}
\setcounter{figure}{0}
\centering
\footnotesize
\caption{Validation performance of the three candidate models on five signs and symptoms spanning a range of prevalence, at each model's selected development iteration (highest development $F1$ across the optimization trajectory). Values in parentheses are 95\% bootstrap confidence intervals. GPT-OSS-20B was selected for the full evaluation.}
\label{tab:modelselect}
\resizebox{\textwidth}{!}{%
\begin{tabular}{llc ccc}
\toprule
Sign/symptom (prevalence) & Model & Iter. & Sens. (95\% CI) & Spec. (95\% CI) & $F1$ (95\% CI)\\
\midrule
Chest pain or discomfort (pain, tightness, or pressure) (12\%) & Gemma~4 31B & 0 & 0.92 (0.75--1.00) & 0.99 (0.97--1.00) & 0.92 (0.78--1.00)\\
 & GPT-OSS-20B & 1 & 1.00 (1.00--1.00) & 0.94 (0.89--0.99) & 0.83 (0.71--0.96)\\
 & Llama~3.1 70B & 5 & 0.75 (0.50--1.00) & 1.00 (1.00--1.00) & 0.86 (0.67--1.00)\\
\addlinespace
Fatigue (14\%) & Gemma~4 31B & 7 & 1.00 (1.00--1.00) & 0.87 (0.79--0.94) & 0.72 (0.61--0.85)\\
 & GPT-OSS-20B & 1 & 1.00 (1.00--1.00) & 0.87 (0.80--0.94) & 0.72 (0.62--0.85)\\
 & Llama~3.1 70B & 7 & 0.79 (0.57--1.00) & 0.99 (0.97--1.00) & 0.85 (0.67--0.97)\\
\addlinespace
Muscle pain (myalgia) (4\%) & Gemma~4 31B & 0 & 1.00 (1.00--1.00) & 0.92 (0.86--0.97) & 0.50 (0.38--0.73)\\
 & GPT-OSS-20B & 2 & 1.00 (1.00--1.00) & 0.93 (0.88--0.97) & 0.53 (0.40--0.73)\\
 & Llama~3.1 70B & 7 & 0.50 (0.00--1.00) & 0.97 (0.93--1.00) & 0.44 (0.00--0.80)\\
\addlinespace
Problems thinking or concentrating (brain fog) (3\%) & Gemma~4 31B & 0 & 1.00 (1.00--1.00) & 0.94 (0.89--0.98) & 0.50 (0.35--0.75)\\
 & GPT-OSS-20B & 4 & 1.00 (1.00--1.00) & 0.95 (0.90--0.99) & 0.55 (0.38--0.86)\\
 & Llama~3.1 70B & 7 & 0.33 (0.00--1.00) & 0.96 (0.92--0.99) & 0.25 (0.00--0.67)\\
\addlinespace
Shortness of breath (22\%) & Gemma~4 31B & 7 & 0.95 (0.86--1.00) & 0.92 (0.86--0.97) & 0.86 (0.76--0.94)\\
 & GPT-OSS-20B & 2 & 1.00 (1.00--1.00) & 0.91 (0.85--0.96) & 0.86 (0.79--0.94)\\
 & Llama~3.1 70B & 7 & 0.91 (0.77--1.00) & 0.87 (0.79--0.94) & 0.77 (0.65--0.86)\\
\midrule
Mean across 5 concepts$^\dagger$ & Gemma~4 31B & -- & 0.97 & 0.93 & 0.70\\
Mean across 5 concepts$^\dagger$ & GPT-OSS-20B & -- & 1.00 & 0.92 & 0.70\\
Mean across 5 concepts$^\dagger$ & Llama~3.1 70B & -- & 0.66 & 0.96 & 0.63\\
\bottomrule
\end{tabular}%
}

\vspace{2pt}
{\raggedright\footnotesize $^\dagger$Means are simple averages of the five point estimates and have no associated CI; per-concept 95\% confidence intervals were computed by bootstrap on the held-out notes.\par}
\end{table}

\begin{table}[H]
\centering
\caption{BERT validation sensitivity by prevalence stratum, across all 72 concepts.}
\label{tab:prevfloor}
\begin{tabular}{lccccc}
\toprule
Prevalence & $n$ & Mean dev.\ positives & BERT Sens. & BERT Spec. & \% concepts, BERT Sens.\,$=0$ \\
\midrule
$\le$2\%   & 37 & 4.6  & 0.03 & 0.98 & 95\% \\
2--5\%     & 17 & 10.5 & 0.19 & 0.95 & 65\% \\
5--10\%    & 8  & 24.2 & 0.56 & 0.79 & 12\% \\
10--22\%   & 10 & 43.7 & 0.78 & 0.68 & 0\% \\ 
\bottomrule
\end{tabular}
\end{table}

\begin{figure}
\centering
\includegraphics[width=\linewidth]{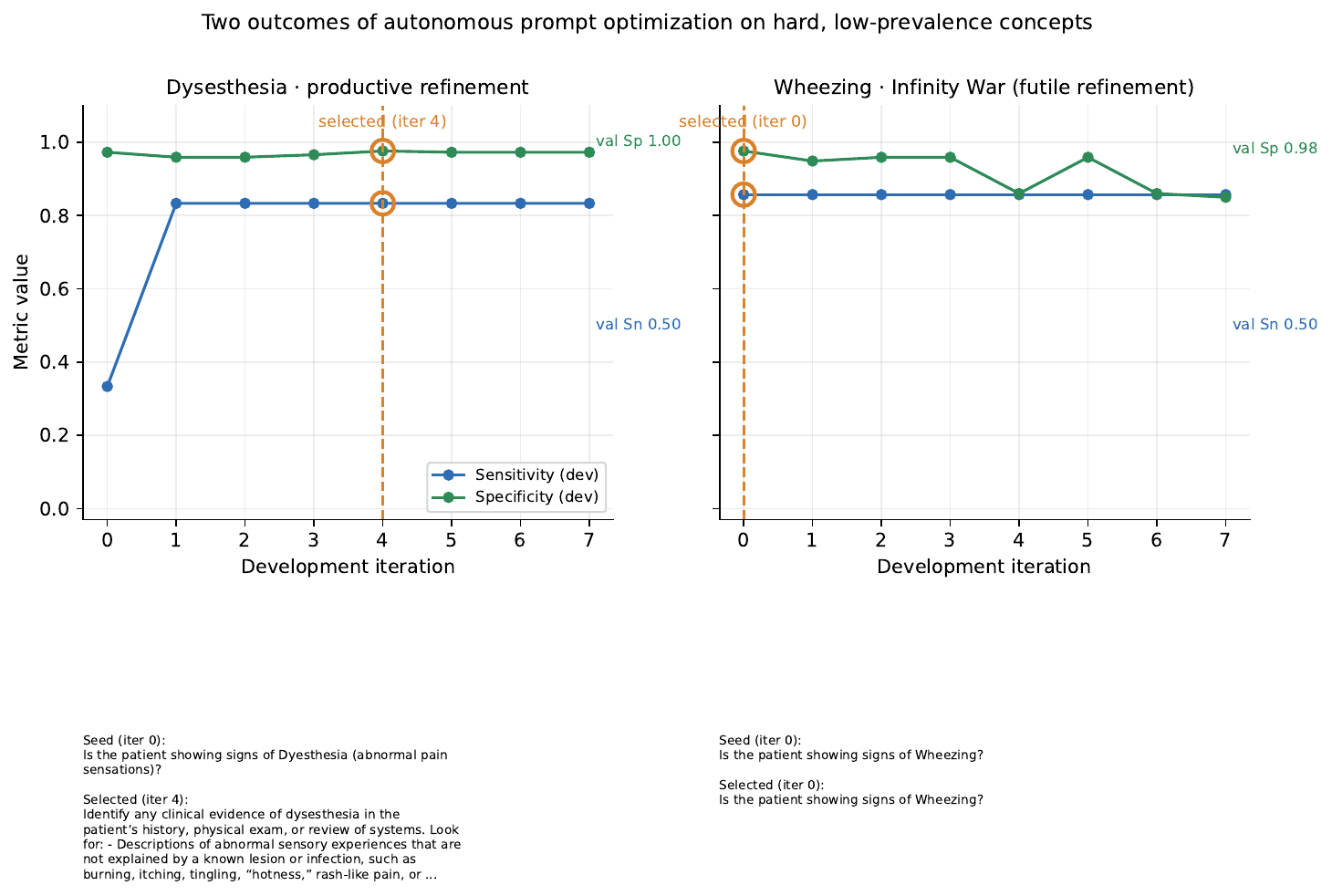}
\caption{Two outcomes of autonomous optimization on low-prevalence concepts for which the lexicon returns a specificity of zero. Development-set sensitivity and specificity are shown across refinement iterations, with the selected iteration circled and the validation result at the right. For dysesthesia (left), refinement recovers sensitivity from 0.33 to 0.83 at a held specificity near 0.97, the agent selects iteration 4, and the selected prompt reaches a validation specificity of 1.00. For wheezing (right), the seed prompt is already strongest, every refinement lowers specificity, the controller backtracks to iteration 0 without changing performance, and the Infinity~War criterion terminates optimization and retains the seed. The seed and selected prompts appear beneath each panel. Both concepts carry a prevalence near 2\%, so the divergent outcomes reflect the optimizer's behavior rather than concept frequency.}
\label{fig:traj}
\end{figure}


\begin{thebibliography}{99}
\bibitem{hripcsak2013} Hripcsak G, Albers DJ. Next-generation phenotyping of electronic health records. \textit{J Am Med Inform Assoc}. 2013;20(1):117--121.
\bibitem{banda2018} Banda JM, Seneviratne M, Hernandez-Boussard T, Shah NH. Advances in electronic phenotyping: from rule-based definitions to machine learning models. \textit{Annu Rev Biomed Data Sci}. 2018;1:53-68.
\bibitem{sheikhalishahi2019} Sheikhalishahi S, Miotto R, Dudley JT, Lavelli A, Rinaldi F, Osmani V. Natural language processing of clinical notes on chronic diseases: systematic review. \textit{JMIR Med Inform}. 2019;7(2):e12239.
\bibitem{wang2018review} Wang Y, Wang L, Rastegar-Mojarad M, et al. Clinical information extraction applications: a literature review. \textit{J Biomed Inform}. 2018;77:34-49.
\bibitem{kreimeyer2017} Kreimeyer K, Foster M, Pandey A, et al. Natural language processing systems for capturing and standardizing unstructured clinical information: a systematic review. \textit{J Biomed Inform}. 2017;73:14-29.
\bibitem{chapman2001negex} Chapman WW, Bridewell W, Hanbury P, Cooper GF, Buchanan BG. A simple algorithm for identifying negated findings and diseases in discharge summaries. \textit{J Biomed Inform}. 2001;34(5):301-310.
\bibitem{savova2010ctakes} Savova GK, Masanz JJ, Ogren PV, et al. Mayo clinical Text Analysis and Knowledge Extraction System (cTAKES): architecture, component evaluation and applications. \textit{J Am Med Inform Assoc}. 2010;17(5):507-513.
\bibitem{soysal2018clamp} Soysal E, Wang J, Jiang M, et al. CLAMP -- a toolkit for efficiently building customized clinical natural language processing pipelines. \textit{J Am Med Inform Assoc}. 2018;25(3):331-336.
\bibitem{uzuner2011i2b2} Uzuner \"{O}, South BR, Shen S, DuVall SL. 2010 i2b2/VA challenge on concepts, assertions, and relations in clinical text. \textit{J Am Med Inform Assoc}. 2011;18(5):552-556.
\bibitem{si2019contextual} Si Y, Wang J, Xu H, Roberts K. Enhancing clinical concept extraction with contextual embeddings. \textit{J Am Med Inform Assoc}. 2019;26(11):1297-1304.
\bibitem{alsentzer2019clinicalbert} Alsentzer E, Murphy J, Boag W, et al. Publicly available clinical BERT embeddings. \textit{arXiv}:1904.03323. 2019.
\bibitem{gu2021pubmedbert} Gu Y, Tinn R, Cheng H, et al. Domain-specific language model pretraining for biomedical natural language processing. \textit{ACM Trans Comput Healthc}. 2022;3(1):1-23.
\bibitem{chen2025benchmark} Chen Q, Hu Y, Peng X, et al. Benchmarking large language models for biomedical natural language processing applications and recommendations. \textit{Nat Commun}. 2025;16(1):3280.
\bibitem{agrawal2022fewshot} Agrawal M, Hegselmann S, Lang H, Kim Y, Sontag D. Large language models are few-shot clinical information extractors. In: \textit{Proc.\ EMNLP}. 2022. \textit{arXiv}:2205.12689.
\bibitem{hu2024prompt} Hu Y, Chen Q, Du J, et al. Improving large language models for clinical named entity recognition via prompt engineering. \textit{J Am Med Inform Assoc}. 2024;31(9):1812-1820.
\bibitem{keloth2024instruction} Keloth VK, Hu Y, Xie Q, et al. Advancing entity recognition in biomedicine via instruction tuning of large language models. \textit{Bioinformatics}. 2024;40(4):btae163.
\bibitem{goel2023annotation} Goel A, Gueta A, Gilon O, et al. LLMs accelerate annotation for medical information extraction. In: \textit{Machine Learning for Health (ML4H)}, PMLR. 2023:82-100.
\bibitem{hsu2025llmie} Hsu E, Roberts K. LLM-IE: a python package for biomedical generative information extraction with large language models. \textit{JAMIA Open}. 2025;8(1):ooaf012.
\bibitem{hu2026ie} Hu Y, Zuo X, Zhou Y, et al. Information extraction from clinical notes: are we ready to switch to large language models? \textit{J Am Med Inform Assoc}. 2026;33(3):553-562.
\bibitem{shin2020autoprompt} Shin T, Razeghi Y, Logan IV RL, Wallace E, Singh S. AutoPrompt: eliciting knowledge from language models with automatically generated prompts. In: \textit{Proc.\ EMNLP}. 2020. \textit{arXiv}:2010.15980.
\bibitem{zhou2022ape} Zhou Y, Muresanu AI, Han Z, et al. Large language models are human-level prompt engineers. In: \textit{Proc.\ ICLR}. 2023. \textit{arXiv}:2211.01910.
\bibitem{yang2023opro} Yang C, Wang X, Lu Y, et al. Large language models as optimizers. In: \textit{Proc.\ ICLR}. 2024. \textit{arXiv}:2309.03409.
\bibitem{khattab2023dspy} Khattab O, Singhvi A, Maheshwari P, et al. DSPy: compiling declarative language model calls into self-improving pipelines. In: \textit{Proc.\ ICLR}. 2024. \textit{arXiv}:2310.03714.
\bibitem{wiest2025local} Wiest IC, Le{\ss}mann M-E, Wolf F, et al. Deidentifying medical documents with local, privacy-preserving large language models: the LLM-anonymizer. \textit{NEJM AI}. 2025;2(4):477-488.
\bibitem{gptoss} OpenAI. gpt-oss-120b \& gpt-oss-20b model card. \textit{arXiv}:2508.10925. 2025.
\bibitem{gemma4} Google DeepMind. Gemma 4 model card [Internet]. 2026 [cited 2026 June 25]. Available from: \url{https://ai.google.dev/gemma/docs/core/model_card_4}
\bibitem{llama31} Grattafiori A, Dubey A, Juahri A, et al. (Llama Team, Meta AI). The Llama 3 herd of models. \textit{arXiv}:2407.21783. 2024.
\bibitem{langgraph} LangChain. LangGraph: building stateful, multi-actor applications with large language models [Internet].  2024 [cited 2026 June 25]. Available from: \url{https://github.com/langchain-ai/langgraph}.
\bibitem{pydantic} Colvin S, Jolibois E, Ramezani H, et al. Pydantic: data validation using Python type hints [Internet].  2024 [cited 2026 June 25]. Available from: \url{https://github.com/pydantic/pydantic}.
\bibitem{estiri_pythia} Cagan C, Fard P, Tian J, Cheng J, Murphy SN, Estiri H. Optimization instability in autonomous agentic workflows for clinical symptom detection. \textit{arXiv}:2602.16037. 2026.
\end{thebibliography}
\end{document}